\documentclass[runningheads,hidelinks]{llncs}
\pdfoutput=1
\usepackage{bm} 
\usepackage{multicol}
\usepackage{amsmath}
\usepackage{hyperref}
\usepackage[inline]{enumitem}
\usepackage[utf8]{inputenc}
\usepackage[english]{babel}
\usepackage{todonotes}
\usepackage{tablefootnote}

\definecolor{blue}{HTML}{1F77B4}
\definecolor{purple}{HTML}{7a00a3}
\definecolor{green}{HTML}{00a35f}
\hypersetup{colorlinks, citecolor=green, urlcolor=purple, linkcolor=purple}
\usepackage{csquotes}
\usepackage[style=numeric, backend=biber, bibencoding=utf8,
    defernumbers=true, maxnames=99, language=english,
]{biblatex}
\makeatletter
\let\blx@rerun@biber\relax
\makeatother
\begin{document}


\title{On an Unknown Ancestor of Burrows' Delta Measure}
\titlerunning{On an Unknown Ancestor of Burrows' Delta Measure}
\author{Petr Plecháč}
\institute{Institute of Czech Literature, Czech Academy of Sciences, Prague, Czech Republic\\
\email{plechac@ucl.cas.cz}}
\maketitle          


\begin{abstract}
This article points out some surprising similarities between a 1944 study by Georgy Udny Yule and modern approaches to authorship attribution.\\[0.3cm]
\textit{Cet article montre l'existence de similitudes surprenantes entre un ouvrage de Georgy Udny Yule  de 1944 et les approches modernes d'attribution d'auteur.
}\end{abstract}

\section{Introduction}

Review articles usually divide the history of using quantitative methods of authorship attribution into two main periods (cf. e.g. \cite{koppel2009, holmes20003}):

\begin{enumerate}
    \item The \textit{univariate approach} era of the 19th and first half of the 20th centuries, which focused mainly on the search for a single textual measure that could distinguish documents written by different authors.

    \item The \textit{multivariate approach} era, which launched with a groundbreaking study by Mosteller and Wallace in 1964.\cite{mosteller1964} Researchers of this era have relied instead on the combined effect of multiple measures and employed multivariate statistical and advanced machine learning methods.
\end{enumerate}

In what follows, I return to a little known chapter from an otherwise influential 1944 study by George Udny Yule. Although Yule’s work is usually seen as an instance of the older univariate approach, in this particular case, he seems to have been quite prescient and treated data in a way that resembles modern multivariate approaches, in particular, John F. Burrows’ well-known Delta measure.\cite{burrows2002,burrows2003} I begin with a brief summary of the Delta principle and then explain the connection with Yule's study.

\section{Burrows' Delta}

In a nutshell, Burrows’ Delta responds as follows to cases where there is a target text of unknown or disputed authorship (t0) and a finite set of texts produced by candidate authors $T = \{t_1, t_2, t_3, \dots, t_m \}$ in a following way:

\begin{enumerate}
\item We extract the $n$ most common words ($w_1, w_2, w_3, \dots, w_n$) in the entire corpus (i.e. $t_0 \cup T$).

\item Each text $t_a \in \{t_0, t_1, t_2, \dots, t_m\}$ is represented as a vector 
\begin{equation}
\mathbf{t_a} = (z_1(t_a), z_2(t_a), \dots, z_n(t_a))
\end{equation}

where $z_i(t_a)$ stands for the $z$-score of relative frequency of a word $w_i$ in the text $t_a$.
\item The stylistic disimilarity between $t_0$ and $t_c \in T$ (the Delta measure $\Delta (\mathbf{t_0}, \mathbf{t_c})$) is calculated as the mean of the absolute differences between the $z$-scores of frequencies of particular words in $t_0$ and $t_c$:

\begin{equation}
\Delta(\mathbf{t_0}, \mathbf{t_c}) =  \sum_{i=1}^{n} \frac{\left|z_i(t_0) - z_i(t_c)\right| }{n}
\end{equation}

\item The target text is attributed to the candidate $c$ which shows the least stylistic dissimilarity from the target, i.e. yields the lowest value of $\Delta(\mathbf{t_0}, \mathbf{t_c})$.
\end{enumerate}

As Shlomo Argamon \cite{argamon2008} has shown, so long as the Delta serves solely as a ranking metric, the division by $n$ (a constant, the number of words analysed) is irrelevant as it in no way affects the ranking of candidate authors. The formula may, thus, be simplified as the Manhattan distance ($L_1$) between vectors $\mathbf{t_0}$ and $\mathbf{t_c}$:

\begin{equation}
\Delta(\mathbf{t_0}, \mathbf{t_c}) \propto L_1(\mathbf{t_0}, \mathbf{t_c}) =  \sum_{i=1}^{n} \left|z_i(t_0) - z_i(t_c)\right|\label{eq:manhattan}
\end{equation}

\noindent Finding the candidate author with the lowest Delta value turns out, then, to mean finding the nearest neighbour according to the Manhattan metric.

There have been several modifications proposed to Burrows’ Delta. Along with the original metric, two such changes have become somewhat standard in authorship recognition studies (cf. e.g. \cite{evert2017}):

\begin{enumerate}
    \item The \textit{Quadratic Delta} ($\Delta_Q$), as proposed in the above-mentioned article by Argamon, which replaces the Manhattan distance with the Euclidean distance ($L_2$)---or more precisely, the Euclidean distance squared:

    \begin{equation}
    \Delta_Q(\mathbf{t_0}, \mathbf{t_c}) = L_2(\mathbf{t_0}, \mathbf{t_c})^2  =  \sum_{i=1}^{n} (z_i(t_0) - z_i(t_c))^2
    \end{equation}
    
    \item The \textit{Cosine Delta} ($\Delta_\angle$) as suggested by Smith and Aldrigde,  \cite{smith2011} which is based on the size of the angle between the vectors (cosine similarity):

    \begin{equation}
    \Delta_\angle(\mathbf{t_0}, \mathbf{t_c}) = 1 - \frac{\sum_{i=1}^{n}z_i(t_0)z_i(t_c)}{\sqrt{\sum_{i=1}^{n}z_i(t_0)^2} {\sqrt{\sum_{i=1}^{n}z_i(t_c)^2}}}
    \end{equation}    
    
\end{enumerate}

\begin{figure}[!htb]
  \centering
      \includegraphics[width=1\textwidth]{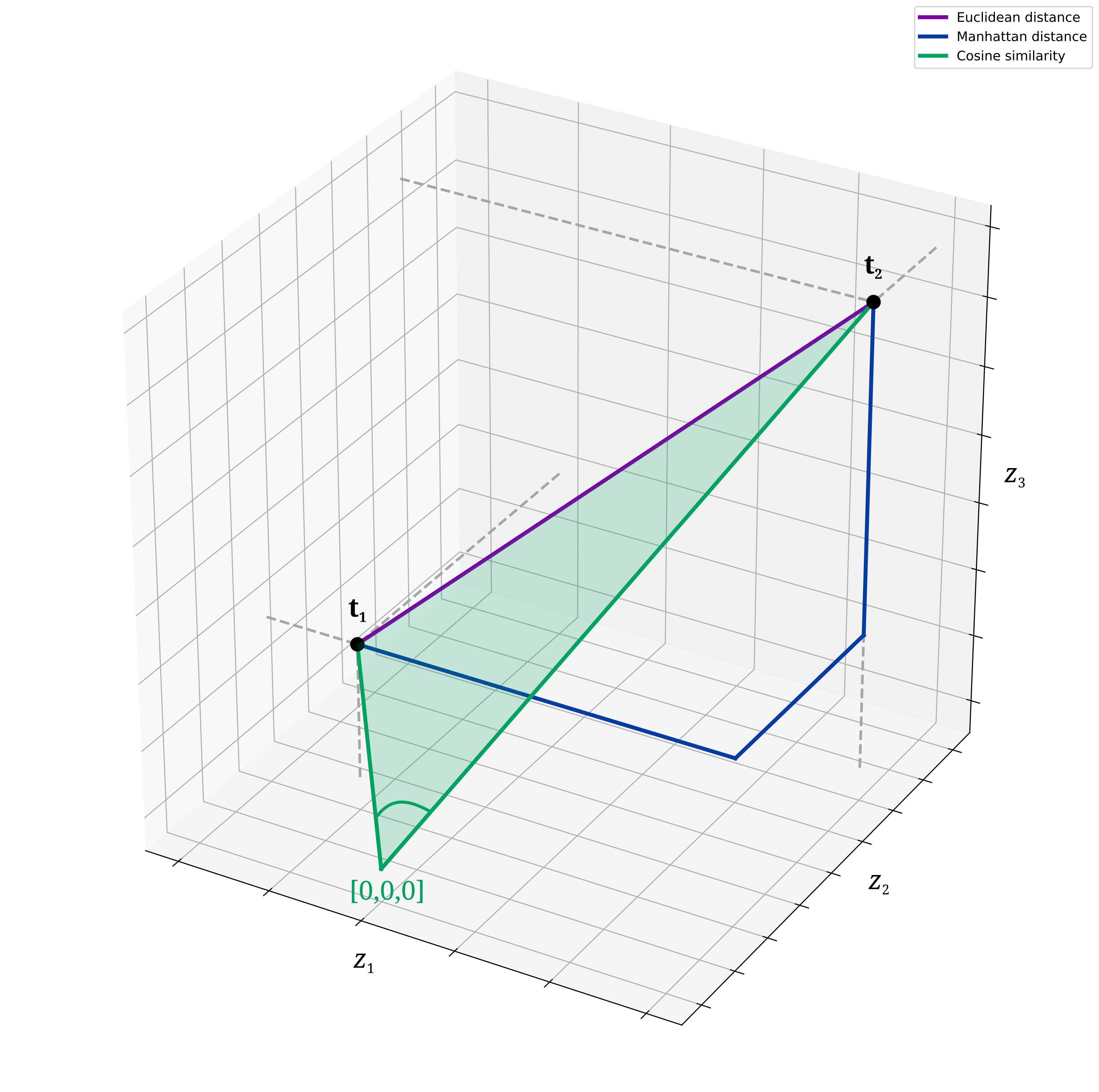}
  \caption{Manhattan distance, Euclidean distance and Cosine similarity between vectors $\mathbf{t_1}$ and $\mathbf{t_2}$}
  \label{fig:dist}
\end{figure}

The Manhattan metric, Euclidean metric and cosine similarity are illustrated in Fig. \ref{fig:dist}.

\section{Yule's Word-Initial Character Method}

Now let us go back several decades. In 1944, George Udny Yule published his book \textit{The Statistical Study of Literary Vocabulary}, which is now widely recognised for introducing Yule’s $K$ (probably the earliest metric of vocabulary richness). This work, however, also contained a little known chapter in which Yule proposed using the frequencies of word-initial characters to discern authorship.

Yule mentions \cite[p.183]{yule1944} that he stumbled on this method quite by accident. His survey of vocabulary richness involved a large card catalogue of the nouns found in particular texts. When two drawers were opened at once---the first containing cards on John Bunyan, the second cards on three essays by Thomas Macaulay---he noticed that the distributions were substantially different. A brief inspection of the drawer for another Macaulay essay showed a card distribution similar to that for the author’s other three essays. This led Yule to consider using frequencies of word-initial characters for the purpose of authorship recognition.

Yule tested this approach with samples from Bunyan's and Macaulay's respective works. In particular, he investigated whether ranking word-initial characters by their frequencies in a sample by author A produced a result more closely resembling the one for the rest of A's data than the one for B's data. In other words, he considered Bunyan's works $t_\textrm{B}$, Macaulay's works $t_\textrm{M}$, a sample $t_0$ extracted from one of them and the 26 letters of the English alphabet $g_1, g_2, ..., g_{26}$. Here the sample and both sets of works are represented by the vectors 

\begin{align*}
\mathbf{t_0} &= (r_1(t_0), r_2(t_0), \dots, r_{26}(t_0))\\
\mathbf{t_B} &= (r_1(t_\textrm{B}), r_2(t_\textrm{B}), \dots, r_{26}(t_\textrm{B}))\\
\mathbf{t_M} &= (r_1(t_\textrm{M}), r_2(t_\textrm{M}), \dots, r_{26}(t_\textrm{M}))\\
\end{align*}

\noindent where $r_i(x)$ stands for the rank of $g_i$ in the frequency-rank distribution of the sample/set of works $x$. 

The goal was to determine which candidate vector $\mathbf{t_c} \in \{\mathbf{t_B},\mathbf{t_M}\}$ was more similar to $\mathbf{t_0}$. Importantly, in pursuing this inquiry, Yule diverged from the then standard stylometric practice of comparing isolated pairs of values. Instead, he aimed to compare the vectors as a whole. He explained this procedure as follows:

\begin{displayquote}
We write down the differences of the ranks in Bunyan sample A from the ranks in the total Bunyan vocabulary, paying no attention to sign; the sum at the foot is a rough measure of the badness of agreement between the sample ranking and for the total of Bunyan vocabulary. In exactly the same way we enter [\ldots] the differences between the sample A ranking and the ranking for the total Macaulay vocabulary, and enter the sum, without regard to sign, at the foot. These respective sums are 10 and 37: we have found that the ranking of the given sample differs much less from that of the Bunyan vocabulary than from that of the Macaulay vocabulary, and are left in practically no doubt that the given sample (if we did not know from which author it had come) should be assigned to Bunyan. \cite[p.190]{yule1944}
\end{displayquote}

What Yule describes as the ``the sum at the foot'' based on the ``differences of the ranks […] paying no attention to sign'' is nothing other than what we now call the \textit{Manhattan distance} between vectors $\mathbf{t_0}$ and $\mathbf{t_c}$:

\begin{equation}
L_1(\mathbf{t_0}, \mathbf{t_c}) = \sum_{i=1}^{n}|r_i(t_0) - r_i(t_c)|
\end{equation}

Interestingly enough, Yule himself noted that this method ``though serving well to bring out the points required, [is] of a very elementary kind and the statistically minded reader may desire to see the results given by more general methods'' \cite[p.191]{yule1944} For this purpose he also offers the Spearman's rank correlation coefficient:\footnote{Yule noted: ``I have followed the usual, but inexact, practice of using this formula even when some of the ranks have been averaged.''\cite[p.191]{yule1944}}

\begin{equation}
\rho(\mathbf{t_0}, \mathbf{t_c}) = 1 - \frac{6 \sum_{i=1}^{n}(r_i(t_0) - r_i(t_c))^2}{n(n^2-1)}\label{eq:spearman}
\end{equation}

Notice that the numerator of the fraction in formula \ref{eq:spearman} equals six times the \textit{Euclidean distance} between $\mathbf{t_0}$ and $\mathbf{t_c}$ squared. Since $n$ is a constant (the number of vector space dimensions = 26), ranking the candidates based on the increasing value of the Spearman's rank correlation coefficient necessarily yields the same result as ranking them based on the decrease in Euclidean distance ($L_2)$:

\begin{equation}
\rho(\mathbf{t_0}, \mathbf{t_c}) = 1 - L_2(\mathbf{t_0}, \mathbf{t_c})^2 \frac{6}{2925}
\end{equation}

\section{Discussion}

Although Yule's feature set (word-initial characters) may seem deficient and arbitrarily chosen from a contemporary perspective, the classification methods he employed were ahead of his time. In particular, his study implicitly introduced the nearest neighbour decision rule decades before its appearance in stylometry and some twenty years before it was established in the science (see \cite{cover1967}).

On the other hand, we should not overstate Yule's contribution. The Manhattan metric is a highly intuitive way of comparing multidimensional data (i.e. as the simple sum of the absolute values of differences) and might be arrived at even without considering its geometrical implications. (Actually, neither Burrows was initially aware of it. As has been shown, it was Shlomo Argamon who connected the dots.)  The relationship between the Euclidean metric and the Spearman's rank correlation coefficient is only indirect. Nevertheless, this study remains noteworthy as an early instance of the multivariate approach in stylometry.


\printbibliography
\end{document}